\documentclass{article} % For LaTeX2e
\usepackage{iclr2027_conference,times}

\usepackage{amsmath,amsfonts,bm}

\def\eqref#1{equation~\ref{#1}}
\def\1{\bm{1}}

\DeclareMathAlphabet{\mathsfit}{\encodingdefault}{\sfdefault}{m}{sl}
\SetMathAlphabet{\mathsfit}{bold}{\encodingdefault}{\sfdefault}{bx}{n}

\usepackage{hyperref}
\usepackage{url}
\usepackage{graphicx}
\usepackage{subcaption}
\usepackage{booktabs}

\newcommand{\dataset}{PrismGaze}
\newcommand{\method}{PrismFusion}
\newcommand{\methodcircle}{PrismFusion-Circle}

\title{Fusing Complementary Multi-view Feature for Screen-based Eye Tracking}

\author{
\begin{tabular}{c}
\textbf{Chang Liu}$^{1}$ \quad
\textbf{Jiaqi Liu}$^{2}$ \quad
\textbf{Chengwen Zhang}$^{2}$ \quad
\textbf{Zhoutong Ye}$^{2}$ \quad
\textbf{Yu Mei}$^{2}$ \\
\textbf{Chun Yu}$^{2,\dagger}$ \quad
\textbf{Yuanchun Shi}$^{2}$ \quad
\textbf{Dong Ma}$^{1}$ \quad
\textbf{Xinjie Shen}$^{3,\dagger}$ \\[4pt]
{\small
$^{1}$University of Cambridge \quad
$^{2}$Tsinghua University \quad
$^{3}$Georgia Institute of Technology
} \\[2pt]
{\small
\texttt{cl2123@cam.ac.uk} \quad
\texttt{xinjie@gatech.edu}
} \quad {\small $^{\dagger}$Corresponding authors}
\end{tabular}
}

\iclrfinalcopy % Uncomment for camera-ready version, but NOT for submission.

\begin{document}

\maketitle

\begin{abstract}
Current multi-view gaze estimation remains limited by existing datasets, insufficient exploitation of complementary cross-view information, and evaluation focused primarily on average gaze error. We address these limitations through a more systematic study of multi-view gaze estimation. First, we introduce \dataset{}, a new dataset with over three million images, capturing continuous head-pose variation for the same gaze targets. Second, we propose \method{}, a multi-view feature fusion framework based on region partitioning, which masks complementary image regions across views during training to encourage effective cross-view information integration. Third, we develop a broader evaluation framework that examines the effects of camera number and placement, target location, and viewing depth.
% , and unseen viewpoints. 
Our experiments show that the primary benefit of multi-view gaze estimation comes from compensating for poorly observed views with cameras providing more favorable viewpoints. \method{} remains robust to changes in viewing depth.
% and unseen viewpoints, while achieving strong calibration-free performance on seen views. 
Together, our dataset, method, and evaluation provide a more comprehensive foundation for studying multi-view gaze estimation in realistic settings.
\end{abstract}

\section{Introduction}
Gaze estimation or eye tracking is important to human--computer interaction and behavioral analysis, giving it great potential for building intelligent environments. Dedicated eye trackers provide accurate measurements, but their specialized sensors and per-user calibration limit unobtrusive deployment at scale~\citep{bao2024unsupervised}. Instead, an increasing number of work now estimate gaze directly from RGB images. They have enabled gaze estimation during laptop use~\citep{zhang2015appearance}, on mobile devices~\citep{krafka2016eye}, and from ordinary video cameras~\citep{park2020eve}. 
However, most existing methods rely on a single camera, making gaze estimation sensitive to camera placement, face or eye occlusion, and distance change between the user and the camera.

Multi-view eye tracking address several limitations of single-view methods by incorporating complementary information from extra views, leading to improved performance~\citep{lian2019multiview, cheng2023dvgaze}. Beyond that, multi-view information has also been exploited to enhance generalization to unseen viewpoints~\citep{hisadome2024rotation, liu2024uvagaze}. With calibrated cameras, multi-view geometry may reduce the need for session-specific calibration. Depth can be recovered from the disparity from views, and optical and visual axis can be estimated with only one calibration point~\citep{shih2004novel}.
% Several labeled multi-view gaze datasets have also been made publicly available~\citep{zhang2020ethxgaze, qin2022learning}. 

Despite this progress, current multi-view eye tracking still face several challenges in data, methodology, and evaluation. From the \textbf{data} perspective, existing studies rely heavily on a small number of datasets: Eth-XGaze and MPII-NV~\citep{zhang2020ethxgaze, qin2022learning}. ETH-XGaze provides diverse camera viewpoints and broad subject coverage, but its public available data mainly consist of normalized face images with users typically sitting on fixed positions, which limits the explicit use of depth data. Furthermore, each gaze target contains only a single image, providing limited variation in head pose and eye movement: a target on the left side of the screen is often associated with both a leftward head pose and a leftward gaze. In real-world use, however, users may look at the same target with very different head poses, such as turning the head to the right while looking left. Missing such variations may reduce robustness in unconstrained settings.
From the \textbf{methodological} perspective, current fusion method may not fully exploit complementary information across views. Although recent studies commonly use Transformer-like architectures for multi-view fusion, attention alone does not guarantee all necessary information from different information are kept and fully utilized. In our preliminary experiments, straightforward Transformer-based fusion showed similar performance on gaze estimation error compared to averaging independent estimation from different views, suggesting that the additional views are not effectively used.
From the \textbf{evaluation} perspective, current evaluation often focus primarily the overall average of gaze estimation error, overlooking some factors that can substantially effect accuracy in practice, such as viewing distance, camera placement, and target position. More comprehensive evaluation is therefore needed to characterize when and where multi-view gaze estimation is effective.

In this work, we address this challenges through a more systematic formulation of multi-view gaze estimation. First, we introduce \dataset{}, a new multi-view gaze dataset containing 24 participants, 288 gaze targets, eight cameras (plus extra two side cameras), and 45 frames per target. Unlike existing datasets with limited head--eye variation for each target, \dataset{} captures continuous variation in head pose across frames. Second, we propose a multi-view feature fusion framework \method{} based on region partitioning. During training, each camera retains only a subset of the original image while the remaining regions are masked, encouraging Transformer to integrate complementary information across views. Finally, we introduce a broader evaluation framework. We study the effects of camera number and placement, target location, and viewing depth.
% , and unseen viewpoints. 
Our results show that the main benefit of \method{} comes from compensating for poorly observed views with cameras that provide more favorable viewpoints. \method{} also remains robust to changes in viewing depth, 
% unseen camera viewpoints, 
while maintaining strong calibration-free performance on seen views.

\section{Related Works}
% \subsection{Single View Eye Tracking}
Single-view eye tracking can be generally categorized into model-based or appearance-based method~\citep{zhang2015appearance, hansen2009eye}. Model-based method constructs geometric eye model. PCCR (Pupil Center Corneal Reflection) is typically used, which creates corneal glints and estimate gaze by mapping the vector between pupil center and glint to the corresponding gaze coordinate on the screen~\citep{merchant1974remote, morimoto2005eye, kim2026hifigaze}. Model-based gaze estimation explicitly reconstructs ocular geometry (e.g., pupil center, iris ellipse) from eye images to infer gaze direction~\citep{chen20083d, kaur2022rethinking}.
The appearance-based method adopts an end-to-end manner that takes an RGB image as an input and outputs either the 3D gaze direction or the exact gaze position on the screen. Many annotated datasets have been collected, whose ground truth can be either the gaze location on screen or the 3d gaze vector~\citep{zhang2015appearance, funes2014eyediap, evdokimov2024weyeds}.

Multi-view camera settings have also been explored for eye tracking. 
% Multiple camera setting allows a better modeling of the eyeball, visual axis, and optical axis~\cite{beymer2003eye, shih2004novel}. However, these model-based methods are sensitive to the shooting angle of cameras therefore typically install on head mounted devices~\cite{bao2024unsupervised}. 
% A number of multi-view datasets have been proposed for appearance based methods.
\citet{lian2019multiview} combines features from three fixed cameras to estimate gaze direction and screen position. Its fixed input structure, however, does not address changes in the number or identity of the available cameras. 
DVGaze introduces structured cross-view correspondence for feature fusion, but is limited to a dual-camera setting~\citep{cheng2023dvgaze}. 
\citet{hisadome2024rotation} use the relative geometry between cameras to guide feature transformation in latent space, enabling gaze estimation to generalize from seen to unseen camera views. 
UVAGaze achieves unsupervised adaptation to a new view by selecting more reliable prediction from the two views to supervise the other \citep{liu2024uvagaze}. 
Many methods have been proposed to fuse the multi-view information. \citet{bao2024unsupervised} uses eye region mask and feature switch between different views to prompt better gaze representation learning. Similarly, \citet{gideon2022unsupervised} and \citet{sun2021cross} leveraged multi-view consistency to learn disentangled gaze representations through latent feature swapping.
Available datasets offer multi-view images with labels. 
EVE records synchronized videos from four user-facing cameras, with gaze label provided by an eye tracker \citep{park2020eve}. 
ETH-XGaze captures broad head-pose and gaze variation with an 18-camera system \citep{zhang2020ethxgaze}. However it provides only cropped face images, which limits the depth estimation. 
MPII-NV~\citep{qin2022learning} reconstructs the 3D facial geometry from the single view dataset MPIIFaceGaze~\citep{zhang2017s}, yet its views are consistent with the ones in ETH-XGaze.

% \citet{zhang2020eth} proposed the ETH-XGaze dataset, in which participants subsequently fixated on 525 targets in front while 18 cameras captured synchronized images of the participant. \citet{park2020towards} proposed EVE dataset, where participants naturally view the screen content while four cameras recorded the video of the participant.

\section{Constructing \dataset{}}

\begin{figure}[t]
    \centering

    \begin{subfigure}[t]{0.48\linewidth}
        \centering
        \includegraphics[width=\linewidth]{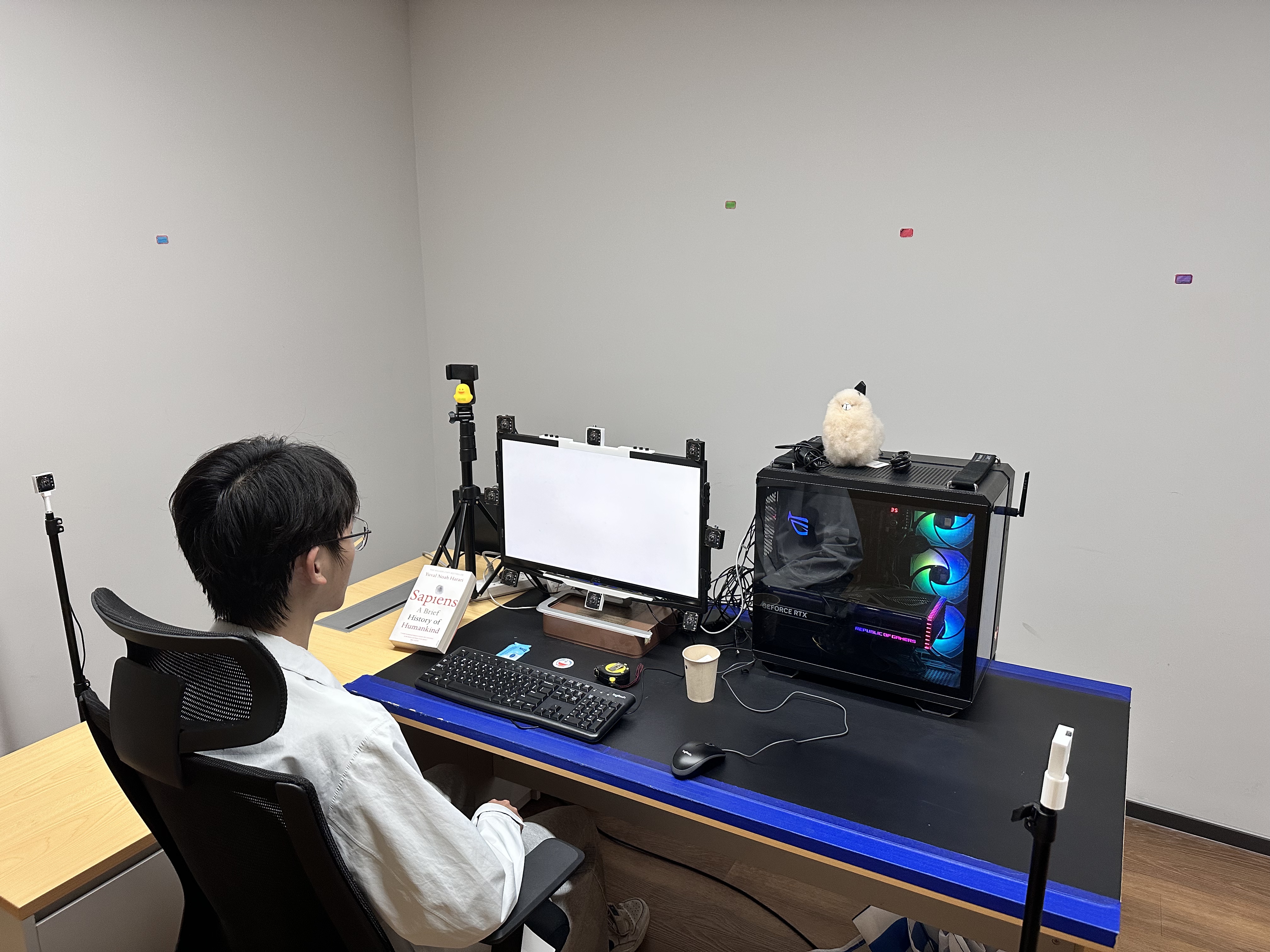}
        \caption{Experimental environment.}
        \label{fig:setup_photo}
    \end{subfigure}
    \hfill
    \begin{subfigure}[t]{0.48\linewidth}
        \centering
        \includegraphics[width=\linewidth]{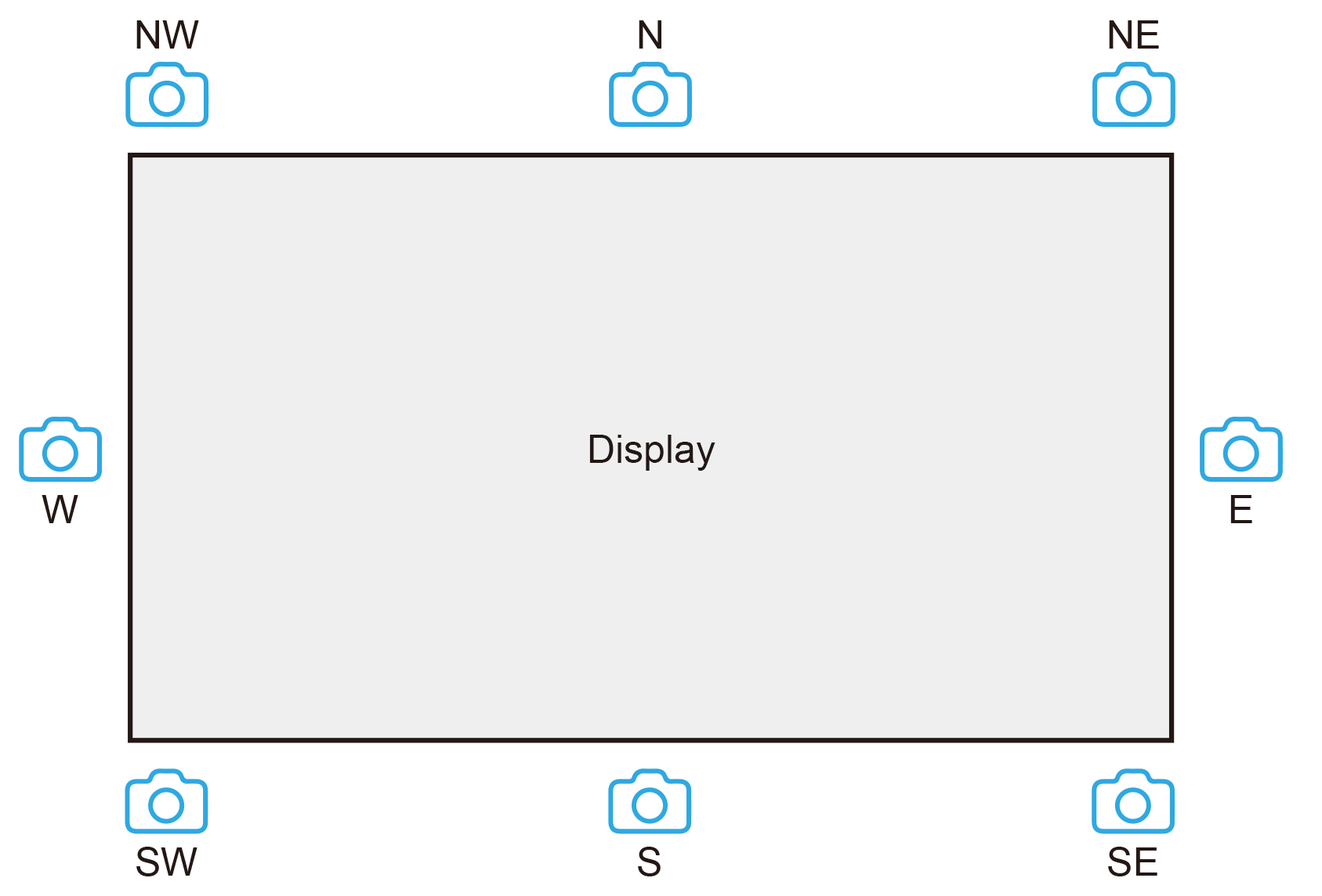}
        \caption{Camera and screen setup.}
        \label{fig:setup_camera}
    \end{subfigure}

    \caption{Experimental setup for \dataset{} data collection. (a) Participants sit approximately 60 cm from the display while ten synchronized RGB cameras and a Tobii Eye Tracker 5 record each fixation trial. (b) Eight cameras are mounted around the display at the NW, N, NE, W, E, SW, S, and SE positions; two additional side-view cameras are placed to the left and right of the participant.}
    \label{fig:setup}
\end{figure}

\subsection{Setup}
The setup for data collection is shown in Fig.~\ref{fig:setup}.
Eight RGB web cameras are rigidly mounted at fixed locations along the display boundary, providing synchronized facial observations from different angles. Besides that, we also mounted two side-view RGB web cameras on the left and right sides of the participant to capture more extreme viewing angles.
All cameras are synchronized and operate at 3840$\times$2160 resolution at 30 FPS with identical settings.

Participants are seated approximately 60cm from a 1920$\times$1080 display (54cm$\times$32.5cm) and are instructed to fixate on predefined targets located either on or outside the screen.
For on-screen targets, we define 288 locations arranged in a regular 24$\times{}$12 grid covering the entire screen area. Each target corresponds to one fixation trial. To collect negative samples, we additional define 12 off-screen targets using physical objects placed on the desk or wall around the participant.

Each fixation trial lasts approximately 1.5 seconds. After fixating on the target, participants are instructed to gently and randomly rotate their head while maintaining their gaze on the target, allowing us to capture the same target under varying gaze and head poses. Each trial is initiated by the participant using keyboard input. During each trial, we record 45 synchronized frames from all camera streams, along with gaze estimates from a Tobii Eye Tracker 5.

The dataset contains 24 participants (13 F, 11M, age 23.15$\pm$3.35). All participants are required to remove glasses during data collection to reduce occlusion.
Each participant completes 288 screen-based trials and 12 out-of-screen trials, resulting in 7,200 trials in total. Since each trial contains 45 synchronized frames recorded from ten cameras, the dataset contains approximately 3.24 million raw images. 

% After preprocessing, each image is further represented by four modalities (full image, face, left eye, and right eye), yielding approximately 12.96 million processed image crops.

\subsection{Annotation}

For eight camera on the boundary, the ground truth of on-screen target is defined as the 2D screen coordinate of the target, and that of off-screen target is defined as a out-screen label.

For the side-view camera, we define the ground-truth as certain area in image. For on-screen targets, we project the predefined 24$\times$12 screen grid into each side-view camera image using a fixed projective transformation. The visible screen region is masked in the input image to prevent the model from directly inferring the target from the displayed content. For off-screen targets, the center of each predefined physical object is used as the corresponding ground-truth location (Details in Sec~\ref{apdx:off_screen_target_ground_truth}).

\subsection{Data Preprocess}
To better support gaze estimation, we crop the original image from on-screen camera for each frame. Four different image are obtained using MediaPipe\footnote{\url{https://developers.google.com/edge/mediapipe/solutions/vision/face_landmarker}} and compressed into 256$\times$256 image: full image, face crop, left eye, right eye. For images in which MediaPipe failed to detect a face, we performed additional detection using 3DDFA~\citep{bulat2017far} and InsightFace\footnote{\url{https://github.com/deepinsight/insightface}}. Images for which all three models failed were manually annotated.

We split the dataset into training, validation, and test sets at a 70\%, 20\%, 10\% ratio using a subject-wise split, ensuring that participants in the three sets were mutually exclusive.

\section{Method}

\begin{figure}
    \centering
    \includegraphics[width=0.95\linewidth]{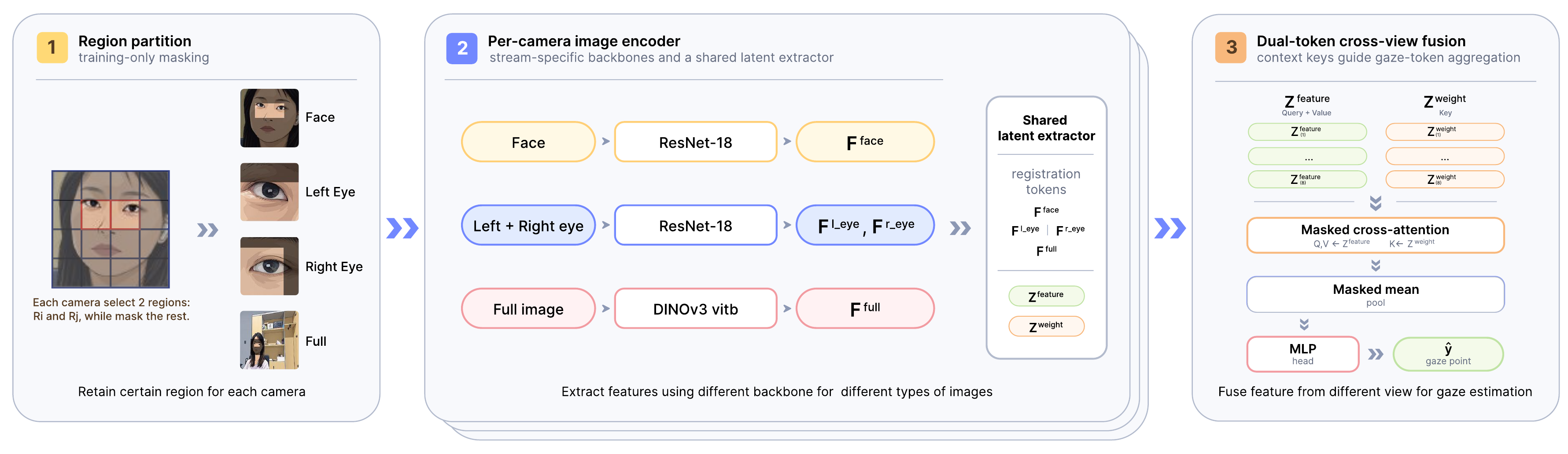}
    \caption{Overview of \method{}. Region Partition distributes cells from a \(4\times4\) face grid across at least two retained cameras and applies the corresponding masks to the full-image, face, and eye crops. A per-view DINO--ResNet encoder produces feature and weight tokens. Cross-view attention uses \(Z^{\mathrm{feature}}\) as queries and values and \(Z^{\mathrm{weight}}\) as keys, followed by mean pooling and an MLP for gaze prediction.}
    \label{fig:method}
\end{figure}

\subsection{Problem Formulation}

Given synchronized observations from \(K\) cameras, our goal is to estimate the two-dimensional screen-space gaze target \(\mathbf{y}\in\mathbb{R}^{2}\). For camera \(k\), the input comprises four aligned image streams:
\[
\mathcal{I}_{k}=
\left\{
I_{k}^{\mathrm{full}},
I_{k}^{\mathrm{face}},
I_{k}^{\mathrm{leye}},
I_{k}^{\mathrm{reye}}
\right\},
\]
corresponding to the full image, face crop, left-eye crop, and right-eye crop. \method{} encodes global and local appearance cues within each view and fuses the camera-wise representations to predict \(\mathbf{y}\).

\subsection{Region Partition}

With complete inputs from every camera, the model can rely on one favorable view or redundant cues across views. Region Partition distributes visible regions across cameras during training and promotes cross-view integration.

For each sample, camera dropout retains at least two cameras. The face crop is divided into a \(4\times4\) grid whose cells are assigned randomly and approximately evenly among the retained cameras. Pixels outside a camera's assigned cells are masked, and the same ownership map is projected to the full-image and eye crops using their crop coordinates. Region Partition is applied only during training; inference uses complete images from all available cameras.

\subsection{Per-View Feature Extraction}

For each camera, the full image is encoded by a DINOv3 ViT-S backbone, the face crop is encoded by a ResNet, and the left- and right-eye crops are processed by another shared ResNet. We project the multi-scale ResNet features into the DINO feature space and convert them into tokens compatible with the DINO patch representation, yielding four feature-token sequences:
\[
\mathbf{F}_{k}^{\mathrm{full}},\quad
\mathbf{F}_{k}^{\mathrm{face}},\quad
\mathbf{F}_{k}^{\mathrm{leye}},\quad
\mathbf{F}_{k}^{\mathrm{reye}}.
\]

The four sequences are concatenated with two learnable query tokens, \(Z^{\mathrm{feature}}\) and \(Z^{\mathrm{weight}}\), and four learnable register tokens. A self-attention encoder mixes the image and latent tokens within each view and outputs two camera-level representations:
\[
\mathbf{z}_{k}^{\mathrm{feature}},\quad
\mathbf{z}_{k}^{\mathrm{weight}}\in\mathbb{R}^{D},
\]
These tokens provide the camera-level representations used for cross-view fusion.

\subsection{Cross-View Fusion}

For the retained camera set \(\mathcal{V}\), the two representations form separate cross-view sequences:
\[
\mathbf{Z}^{\mathrm{feature}}=
\left[\mathbf{z}_{k}^{\mathrm{feature}}\right]_{k\in\mathcal{V}},
\qquad
\mathbf{Z}^{\mathrm{weight}}=
\left[\mathbf{z}_{k}^{\mathrm{weight}}\right]_{k\in\mathcal{V}}.
\]
We employ a stack of cross-view attention blocks in which \(\mathbf{Z}^{\mathrm{feature}}\) provides the queries and values, while \(\mathbf{Z}^{\mathrm{weight}}\) provides the keys. Let \(\mathbf{H}^{(0)}=\mathbf{Z}^{\mathrm{feature}}\); each fusion block updates the camera-wise representation as
\[
\mathbf{H}^{(\ell+1)}
=
\mathrm{FFN}\!\left(
\mathbf{H}^{(\ell)}
+
\mathrm{Attn}\!\left(
\mathbf{H}^{(\ell)},
\mathbf{Z}^{\mathrm{weight}},
\mathbf{H}^{(\ell)}
\right)
\right),
\]
% where invalid cameras are excluded by an attention mask. The final camera-wise features are mean-pooled over valid cameras:
% \[
% \mathbf{h}
% =
% \frac{1}{|\mathcal{V}|}
% \sum_{k\in\mathcal{V}}
% \mathbf{H}_{k}^{(L)}.
% \]
% Finally, an MLP maps \(\mathbf{h}\) to the gaze prediction:
% \[
% \hat{\mathbf{y}}=\mathrm{MLP}(\mathbf{h}).
% \]

The camera-wise features are mean-pooled over retained views and mapped by an MLP to the gaze prediction \(\hat{\mathbf{y}}\). Training minimizes the Euclidean distance between the predicted and ground-truth gaze locations over \(\mathcal{D}\):
\[
\mathcal{L}_{\mathrm{gaze}}
=
\frac{1}{|\mathcal{D}|}
\sum_{(\mathcal{I},\mathbf{y})\in\mathcal{D}}
\left\|
\hat{\mathbf{y}}-\mathbf{y}
\right\|_{2}.
\]

\section{Evaluation}

\subsection{Experimental Setup}

We evaluate camera-count effects on both \dataset{} and ETH-XGaze, camera placement and target location on \dataset{}, and viewing distance on an additional real-world dataset. \dataset{} uses a 70\%/20\%/10\% subject-wise split with no participant overlap. For ETH-XGaze, 10 participants from the official training set are randomly selected for validation, the remaining 70 are used for training, and the official test set is used for evaluation. The checkpoint with the lowest validation error is evaluated on the corresponding test set. Unless otherwise stated, we report mean gaze-prediction error over all evaluation samples.

\subsection{Performance on \dataset{} and ETH-XGaze}

\begin{table}[t]
    \centering
    \footnotesize
    \begin{tabular}{llcc}
            \toprule
            \textbf{Method} &
            \textbf{Implementation} &
            \textbf{\# Cams} &
            \textbf{Error (cm)} \\
            \midrule
            \method{} & Ours & 8 & \textbf{0.94} \\
            \citet{lian2019multiview} & Reimplemented & 8 & 1.46 \\
            DVGaze~\citep{cheng2023dvgaze} (8-view) & Extended from official code & 8 & 1.33 \\
            GazeTR~\citet{cheng2022gaze} (8-view) & Extended from official code & 8 & 2.49 \\
            \cmidrule(lr){1-4}
            \method{} (2-view) & Ours & 2 & \textbf{1.72} \\
            Rot-MVGaze~\citep{hisadome2024rotation} & Official code & 2 & 5.21 \\
            UVAGaze~\citep{liu2024uvagaze} & Official code & 2 & 7.27 \\
            DVGaze~\citep{cheng2023dvgaze} & Official code & 2 & 2.42 \\
            \bottomrule
    \end{tabular}
    \caption{Comparison on \dataset{}. We report mean two-dimensional screen-space error in centimeters (lower is better) for the eight-view setting and the E/W two-view setting. Bold values mark the best result within each setting.}
    \label{tab:method_comparison}
\end{table}

\begin{table}[t]
    \centering
    \footnotesize
    \setlength{\tabcolsep}{4pt}
    \begin{tabular}{lccc}
        \toprule
        \multicolumn{1}{c}{\textbf{Method}} & \multicolumn{1}{c}{\textbf{\# Cameras}} & \textbf{Error Within (°)} & \textbf{Error Person Specific (°)} \\ \hline
        % \method{} & 10 & 2.55 & 2.97 \\
        \method{} (10 views) & 10 & 2.89 & 3.25 \\
        \method{} (2 views) & 2 & 3.14 & 3.00 \\
        \method{} (all views) & 18 & 3.49 & 3.80 \\
        Rot-MVGaze~\citep{hisadome2024rotation} & 2 & \multicolumn{2}{c}{3.5$^{\dagger}$} \\
        UVAGaze~\citep{liu2024uvagaze} & 2 & \multicolumn{2}{c}{3.27} \\
        DVGaze~\citep{cheng2023dvgaze} & 2 & \multicolumn{2}{c}{2.46$^{\ddagger}$} \\
        \bottomrule
        \end{tabular}
    \caption{Angular gaze error on ETH-XGaze in degrees (lower is better). \method{} is evaluated with the 10 inner-ring cameras, nine two-view camera pairs, and all 18 cameras. Results for prior methods are quoted from their respective papers; as their evaluation protocols differ from ours, their values are placed between the two error columns rather than assigned to either. $^{\dagger}$Evaluated with 4-fold subject cross-validation. $^{\ddagger}$Potential data leakage identified through our inspection of the publicly available source code.}
    \label{tab:method_comparison_eth}
\end{table}

% UVAGaze
% | Pair | 两个 camera      |
% | ---- | -------------- |
% | 1    | cam 0 + cam 9  |
% | 2    | cam 1 + cam 10 |
% | 3    | cam 2 + cam 11 |
% | 4    | cam 3 + cam 12 |
% | 5    | cam 4 + cam 13 |
% | 6    | cam 5 + cam 14 |
% | 7    | cam 6 + cam 15 |
% | 8    | cam 7 + cam 16 |
% | 9    | cam 8 + cam 17 |

For methods with public code, we adapt the released implementations to match our input and output formats. \method{} has the lowest mean error in both the eight-view and E/W two-view comparisons in Table~\ref{tab:method_comparison}.

Table~\ref{tab:method_comparison_eth} reports published ETH-XGaze results together with \method{}. The all-view setting uses 18 cameras, the 10-view setting uses the inner ring, and the two-view setting follows the nine camera pairs used by UVAGaze~\citep{liu2024uvagaze}: $(0,9)$, $(1,10)$, $(2,11)$, $(3,12)$, $(4,13)$, $(5,14)$, $(6,15)$, $(7,16)$, and $(8,17)$. The 18-view setting has higher error than the 10-view setting under both metrics, showing that using all available views does not necessarily improve performance.

\subsection{Ablation Studies}

Removing Region Partition while keeping the backbone configuration fixed increases error from 0.94 cm to 1.63 cm (Table~\ref{tab:ablation}). Changing the ResNet assignments produces errors from 1.38 cm to 1.82 cm, while using DINO backbones for all inputs gives substantially higher errors. Region Partition and the mixed backbone design form the final configuration.

\begin{table}[!htbp]
    \centering
    \footnotesize
    \setlength{\tabcolsep}{3pt}
    \begin{tabular}{cccllllc}
        \toprule
        \textbf{Index} &
        \textbf{Region Partition} &
        \textbf{DINO Finetune} &
        \textbf{Full} &
        \textbf{Face} &
        \textbf{Left Eye} &
        \textbf{Right Eye} &
        \textbf{Error (cm)} \\
        \midrule
        0 & Yes & No  & DINO         & ResNet$^{0}$ & ResNet$^{1}$ & ResNet$^{1}$ & \textbf{0.94} \\
        \midrule
        1 & No  & No  & DINO         & ResNet$^{0}$ & ResNet$^{1}$ & ResNet$^{1}$ & 1.63 \\
        2 & Yes & No  & DINO         & ResNet$^{0}$ & ResNet$^{1}$ & ResNet$^{2}$ & 1.47 \\
        3 & Yes & No  & ResNet$^{0}$ & ResNet$^{1}$ & ResNet$^{2}$ & ResNet$^{2}$ & 1.38 \\
        4 & Yes & No  & ResNet$^{0}$ & ResNet$^{1}$ & ResNet$^{2}$ & ResNet$^{3}$ & 1.82 \\
        5 & Yes & No  & DINO$^{0}$   & DINO$^{1}$   & DINO$^{2}$   & DINO$^{2}$   & 6.05 \\
        6 & Yes & No  & DINO$^{0}$   & DINO$^{1}$   & DINO$^{2}$   & DINO$^{3}$   & 6.35 \\
        7 & Yes & Yes & DINO$^{0}$   & DINO$^{1}$   & DINO$^{2}$   & DINO$^{2}$   & 3.58 \\
        8 & Yes & Yes & DINO$^{0}$   & DINO$^{1}$   & DINO$^{2}$   & DINO$^{3}$   & 6.87 \\
        \bottomrule
    \end{tabular}
    \caption{Ablation of Region Partition, DINO fine-tuning, and backbone assignments on \dataset{}. We report mean screen-space error in centimeters (lower is better); the bold row is the final \method{} configuration. Different values of $i$ in $Backbone^{i}$ indicate different backbones.}
    \label{tab:ablation}
\end{table}

\subsection{Effect of the Camera Count}
\label{subsec:effect_of_camera_count}

On \dataset{}, \method{} is evaluated with one to eight available cameras. All combinations are used for the one- and eight-camera settings; for each camera count from two to seven, eight valid subsets are sampled randomly.

% For larger subsets, the results characterize the performance obtained as the system receives increasingly complete multi-view observations.
ETH-XGaze contains 10 inner-ring cameras and 8 outer-ring cameras. One sampling scheme draws up to 10 cameras from the inner ring and then adds outer-ring cameras; the other draws up to 8 cameras from the outer ring and then adds inner-ring cameras. Three configurations are sampled randomly for each camera count in each scheme.

% Then, we evaluate that on ETH-XGaze. For camera counts up to eight, we consider three sampling protocols: subsets drawn from the ten inner camera ring, subsets drawn from the eight outer ring, and mixed subsets containing four inner-ring and four outer-ring cameras. For camera counts above eight, subsets are sampled randomly from all available cameras.
% To further isolate the value of dense central viewpoints, we exhaustively evaluate the inner-ring settings with \(8\), \(9\), and \(10\) cameras. These results reveal whether performance gains arise simply from camera quantity or also from the geometric distribution of viewpoints.
% This experiment answers whether the proposed fusion architecture can effectively exploit additional views, especially when each individual view contains only partial or unfavorable facial evidence.

\begin{figure}[t]
    \centering

    \begin{subfigure}[t]{0.32\linewidth}
        \centering
        \includegraphics[width=\linewidth]{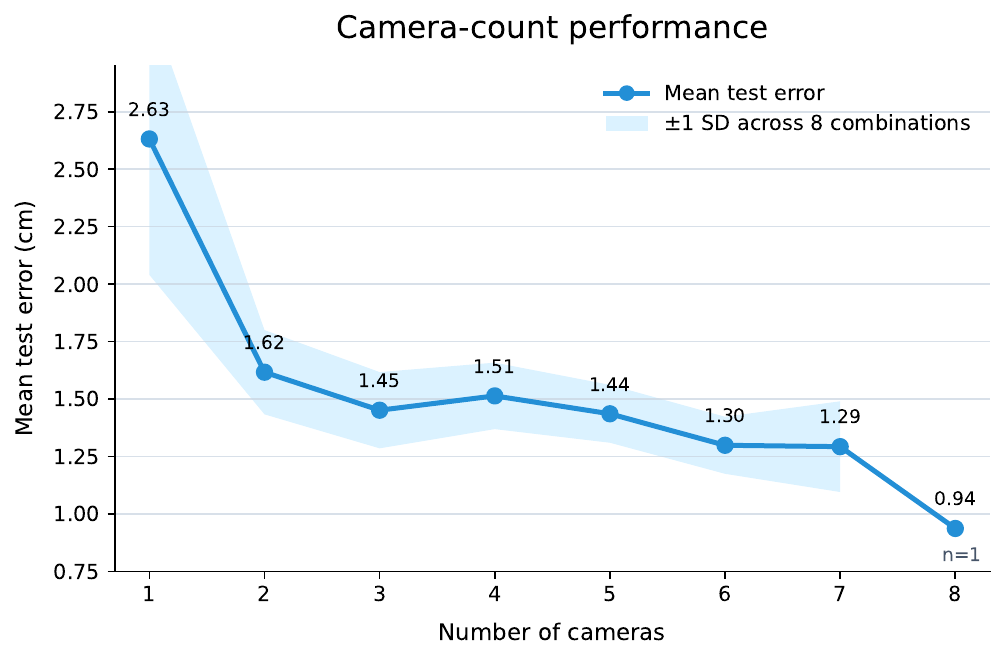}
        \caption{\dataset{}}
        \label{fig:camera_count_prism}
    \end{subfigure}
    \hfill
    \begin{subfigure}[t]{0.32\linewidth}
        \centering
        \includegraphics[width=\linewidth]{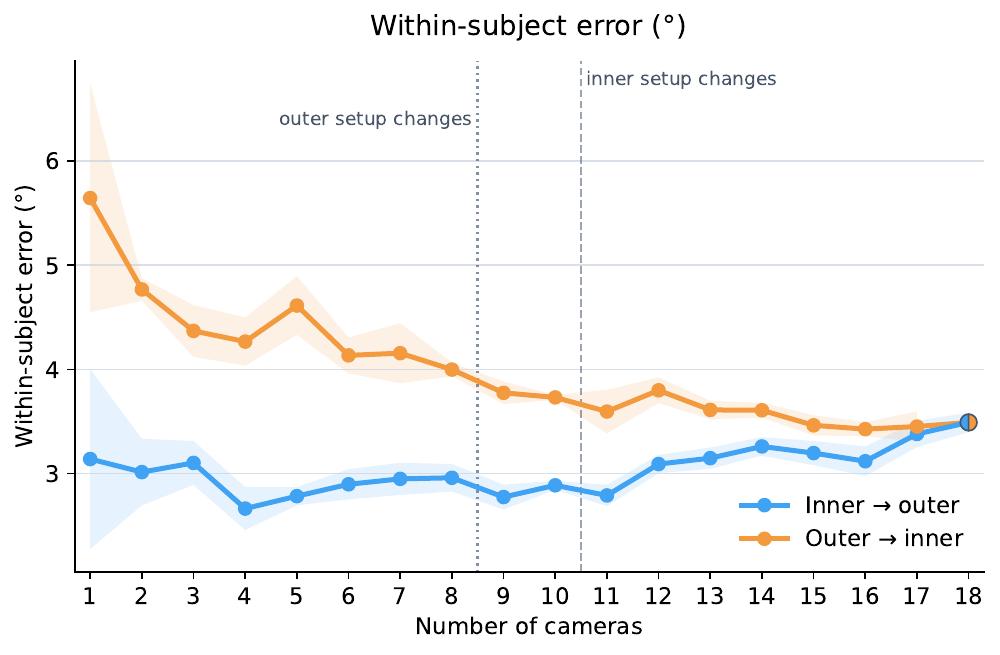}
        \caption{ETH-XGaze Within Subject}
        \label{fig:camera_count_xgaze_within}
    \end{subfigure}
    \hfill
    \begin{subfigure}[t]{0.32\linewidth}
        \centering
        \includegraphics[width=\linewidth]{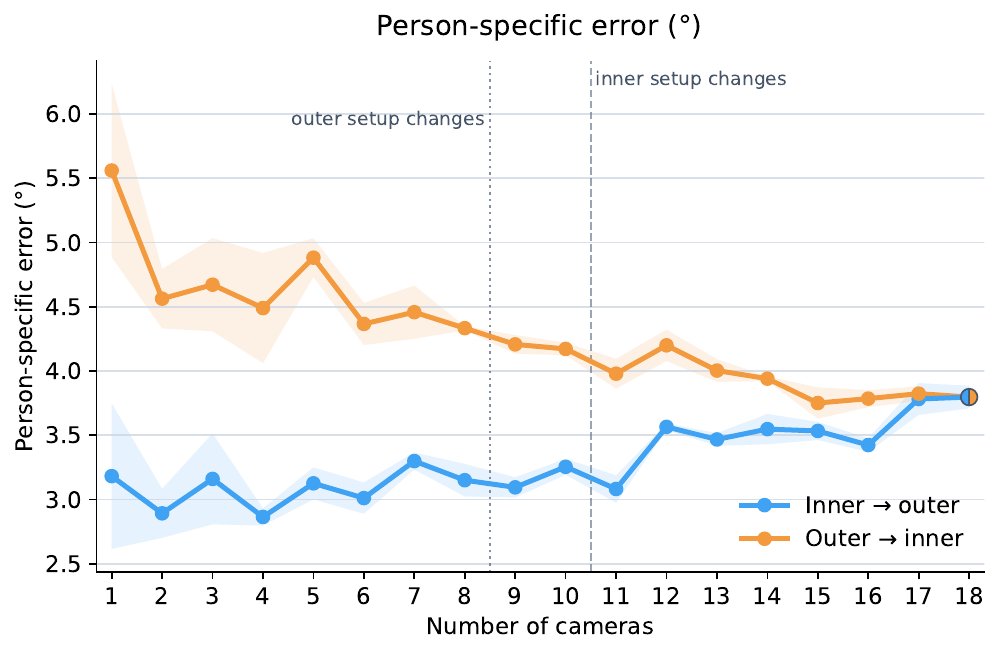}
        \caption{ETH-XGaze Person Specific}
        \label{fig:camera_count_xgaze_person}
    \end{subfigure}
    \caption{Effect of camera count and composition. (a) Mean test error on \dataset{} across camera subsets; shading denotes one standard deviation. (b--c) ETH-XGaze errors under inner-to-outer and outer-to-inner camera-sampling orders for the within-subject and person-specific protocols. Lower values indicate better performance.}
    \label{fig:camera_count}
\end{figure}

On \dataset{}, error generally decreases as more cameras become available, although the trend is not monotonic (Fig.~\ref{fig:camera_count}). On ETH-XGaze, inner-ring subsets perform better at small camera counts, while adding outer-ring cameras does not consistently reduce error. Conversely, outer-ring subsets improve after inner-ring cameras are added. Camera composition therefore matters alongside camera count.

\subsection{Effect of Camera Placement}
% The number of cameras alone does not fully determine multi-view gaze estimation performance. 
We analyze camera placement using single-camera results and two- and three-camera configurations drawn from the eight fixed positions; Sec.~\ref{apdx:camera_configuration_analysis} gives the configuration details.

% The number of cameras alone does not fully determine the quality of multi-view gaze estimation. We therefore evaluate how different spatial configurations of certain camera number affect performance.

% This analysis is conducted on the in-house dataset. We select representative camera subsets of sizes one to six, including balanced configurations, one-sided configurations, corner-dominated configurations, edge-dominated configurations, and configurations spanning opposite sides of the screen. The seven-camera setting is covered by the random-subset experiment above.

% For three-camera subsets, we additionally include configurations in which all cameras lie on the same side of the setup. For four-camera subsets, we consider configurations that densely cover one corner while adding a camera from another corner, as well as configurations that cover one full edge and include one camera from another edge. Similar structured configurations are evaluated for five- and six-camera subsets, including two fully covered edges and two fully covered corners.

% By comparing configurations with the same number of cameras, this experiment quantifies the importance of viewpoint diversity and spatial coverage.

Single-camera error ranges from 1.76 cm for Camera E to 3.50 cm for Camera NE (Table~\ref{tab:spatial_error}(a)), showing substantial variation across camera positions.

To test whether single-camera quality carries over to camera pairs, we rank the eight positions by single-camera error and compute the mean rank of each pair. This mean rank correlates with the ranking of the final two-camera error (Spearman $\rho=0.553$, position-label permutation $p<0.05$): pairs containing stronger individual cameras tend to have lower error.

Spatial spread is measured by inter-camera distance for camera pairs and by convex-hull area for three-camera configurations, with zero area assigned to collinear positions. Neither measure is significantly associated with test error (Spearman $p>0.05$ in both cases).
% (Spearman $\rho=-0.104$, permutation test $p=0.595$), 
% and the linear association is similarly weak (Pearson $r=-0.076$, $p=0.699$, $R^2=0.006$). 
% (Spearman $\rho=-0.160$, permutation test $p=0.530$; Pearson $r=-0.089$, $p=0.725$). 

These results show that camera quality is more informative than spatial spread alone when selecting camera combinations.

% Thus, the results do not support either the hypothesis that more widely separated cameras consistently reduce error or that more closely clustered cameras consistently improve performance. 

\subsection{Error Analysis Across Target Locations}

The display is divided into nine regions corresponding to the eight camera positions and the screen center. Each target is assigned to its nearest camera or the center point; Table~\ref{tab:spatial_error}(b) reports the resulting errors.

Screen region has a significant effect on prediction error (one-way ANOVA, $F=19.40$, $p<0.001$). Bonferroni-corrected Welch's t-tests show that Center has lower error than W, SW, E, S, SE, and NE, but does not differ significantly from NW or N. NW has lower error than SW, S, SE, and NE; E has lower error than S and SE; and N has lower error than SW, S, SE, and NE. The lowest mean errors occur at Center, NW, and N, while S, SE, and SW have the highest errors.

\begin{table}[t]
    \centering
    \footnotesize
    \setlength{\tabcolsep}{5pt}
    \begin{tabular}{cc@{\hspace{1.0cm}}lcc}
        \toprule
        \multicolumn{2}{c}{\textbf{(a) Single-camera error}} &
        \multicolumn{3}{c}{\textbf{(b) Error across target regions}} \\
        \cmidrule(lr){1-2}\cmidrule(lr){3-5}
        \textbf{Position} & \textbf{Error (cm)} &
        \textbf{Region} & \textbf{Targets} & \textbf{Error (cm)} \\
        \midrule
        E  & 1.76 & Center & 72  & 0.73 \\
        SE & 2.00 & NW     & 18  & 0.78 \\
        S  & 2.37 & N      & 36  & 0.78 \\
        W  & 2.51 & E      & 36  & 0.89 \\
        NW & 2.72 & W      & 36  & 0.90 \\
        SW & 3.07 & NE     & 18  & 1.02 \\
        N  & 3.14 & S      & 36  & 1.07 \\
        NE & 3.50 & SE     & 18  & 1.16 \\
           &      & SW     & 18  & 1.44 \\
        \bottomrule
    \end{tabular}
    \caption{Spatial error analysis on \dataset{}. (a) Single-camera error by camera position. (b) Multi-view gaze-estimation error by target region, reported in both pixels and centimeters. Rows in each panel are ordered by increasing centimeter error; lower values indicate better performance.}
    \label{tab:spatial_error}
\end{table}

\begin{figure}[t]
    \centering

    \begin{subfigure}[t]{0.48\linewidth}
        \centering
        \includegraphics[width=\linewidth]{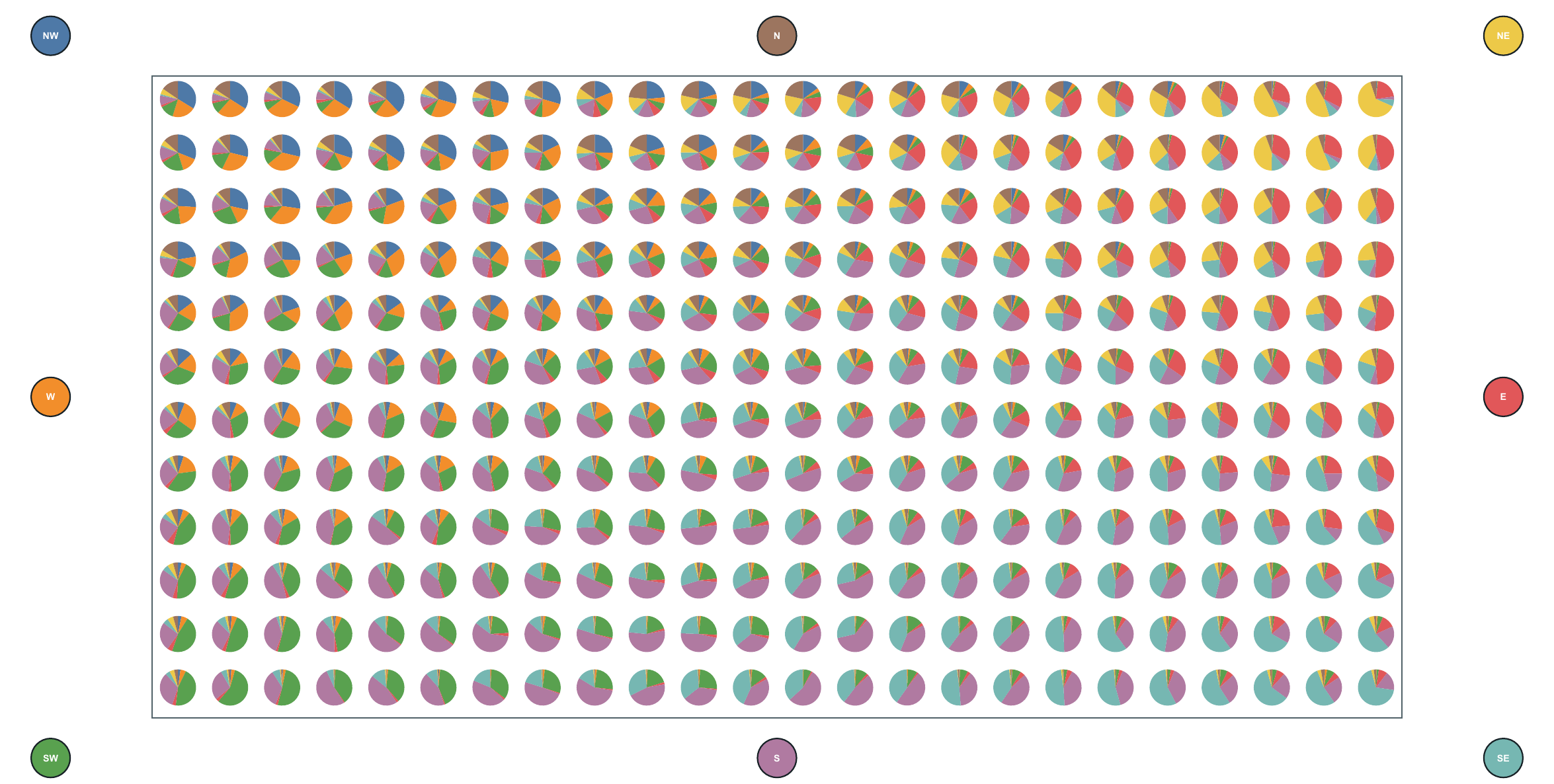}
        \caption{With Region Partition.}
        \label{fig:attn_dstrb_with_rp}
    \end{subfigure}
    \hfill
    \begin{subfigure}[t]{0.48\linewidth}
        \centering
        \includegraphics[width=\linewidth]{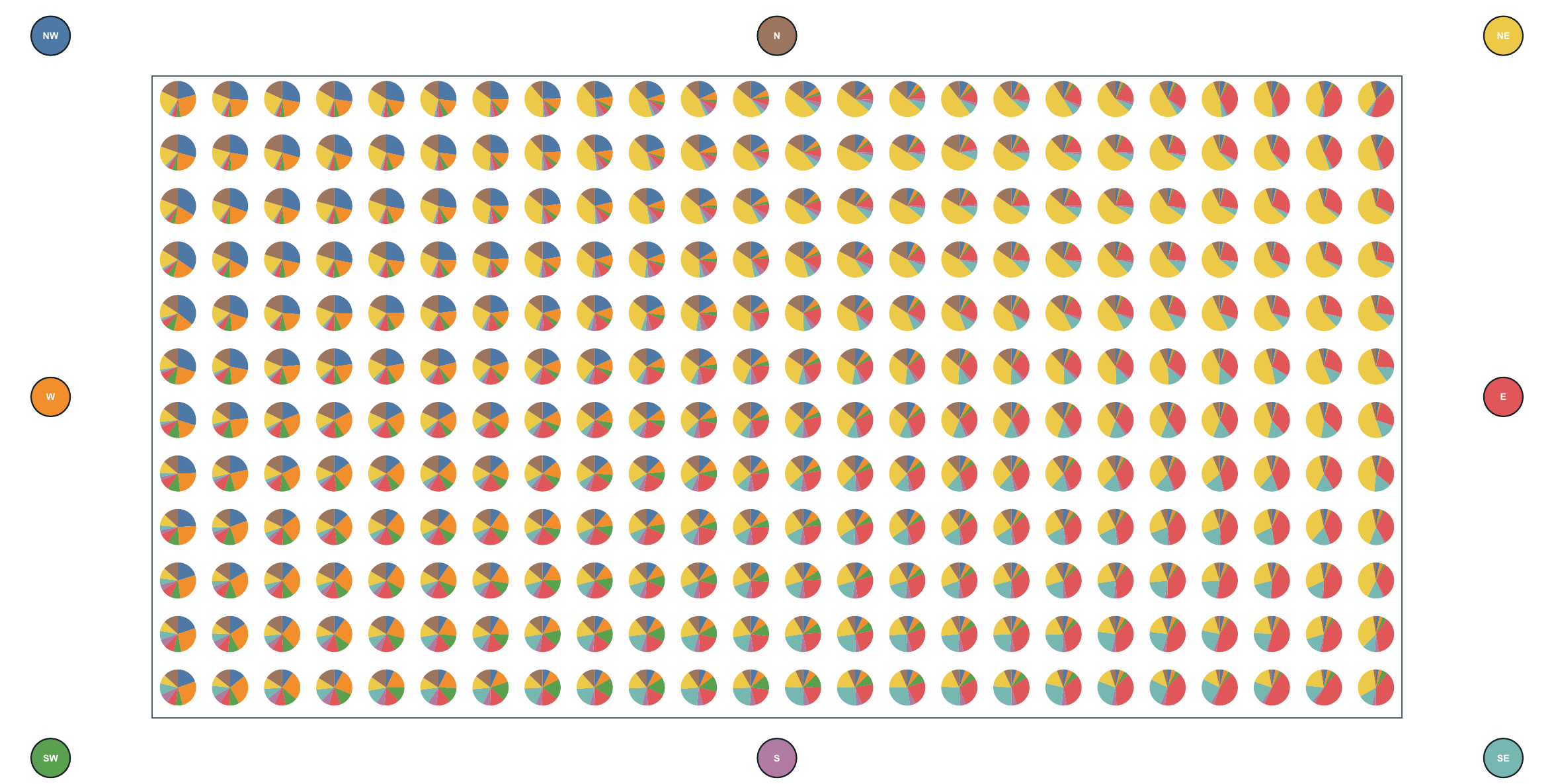}
        \caption{Without Region Partition (ablation 1 in Table~\ref{tab:ablation}).}
        \label{fig:attn_dstrb_wo_rp}
    \end{subfigure}

    \caption{Target-wise cross-view attention with and without Region Partition. Each pie chart corresponds to one gaze target and shows the normalized attention mass assigned to the eight camera keys, with colors indicating camera position. Region Partition produces stronger spatial variation in camera attention across the screen.}
    \label{fig:attn_dstrb}
\end{figure}

Figure~\ref{fig:attn_dstrb} compares camera-level attention mass with and without Region Partition. In cross-attention, \(Z^{\mathrm{feature}}\) provides the queries and values and \(Z^{\mathrm{weight}}\) provides the keys. Attention distributions over key cameras are similar across query cameras, so we aggregate across queries using the number of valid samples as weights, average across layers, and normalize over key cameras.

Attention mass is negatively correlated with target-to-camera distance, with a stronger correlation under Region Partition (\(\rho=-0.6763\)) than without it (\(\rho=-0.3730\)). Thus, cameras closer to a target tend to receive more attention.

Attention mass is also negatively correlated with the corresponding single-camera error (\(\rho=-0.345\), position-label permutation \(p<0.001\)). Region Partition produces more spatially structured attention, assigning greater attention to cameras that are closer to the target and achieve lower single-view error.

\subsection{Error Analysis in Real-world Scenario}

% Using real-device data collected at different subject-to-screen distances, we evaluate the relationship between gaze error and viewing depth. This experiment is conducted under the full eight-camera configuration to avoid conflating depth effects with missing-view effects.

% The results show whether the learned representation generalizes across changes in apparent face scale, perspective, and head-to-camera distance.

The viewing-distance evaluation uses a separate group of 12 participants (6 female and 6 male; age $23.7\pm3.75$ years). Participants gaze at six peripheral calibration points and 44 points sampled independently for each participant on the same display as the main study. Data are collected at 60, 80, and 100 cm.

\methodcircle{} provides a geometric refinement by predicting the distance from the gaze point to each camera rather than \(\hat{\mathbf{y}}\) directly. The predicted distances define circles centered at the eight camera locations, and nonlinear least squares recovers the point that best fits them. We compare \method{}, \methodcircle{}, and their averaged prediction. For the two-view setting, we report the camera pair with the lowest mean error across the three distances.

Calibration fits an affine transformation for each participant and distance using the six peripheral targets and applies it to the remaining 44 targets.

At 60 and 80 cm, Tobii has lower error than all uncalibrated methods and most calibrated variants in Table~\ref{tab:distance_evaluation}. Calibrated two-view \method{} does not differ significantly from Tobii at 80 cm.

At 100 cm, calibrated \methodcircle{} does not differ significantly from Tobii, whereas the calibrated averaged prediction and calibrated two-view \method{} have lower error than Tobii (2.80 and 2.49 cm, respectively; $p<0.01$). Calibration reduces the error of all \method{} variants at each distance.

\begin{table}[t]
    \centering
    \footnotesize
    \begin{tabular}{lccc}
        \toprule
        \textbf{Method} &
        \textbf{60 cm} &
        \textbf{80 cm} &
        \textbf{100 cm}\\
        \midrule
        Tobii Eye Tracker & \textbf{1.38}$^{\ast\ast\ast}$  & \textbf{1.80}$^{\ast\ast\ast}$  & 3.42\\
        \midrule
        \method{} & 9.62 & 10.98 & 12.31 \\
        \methodcircle{} & 5.33 & 6.32 & 7.60 \\
        Average of two & 5.58 & 6.47 & 8.03 \\
        \method{} (2 views) & 6.11 & 5.89 & 7.93 \\
        \midrule
        \method{} calibrated & 2.88 & 3.06 & 3.57 \\
        \methodcircle{} calibrated & 2.43  & 2.76  & \textbf{3.48}$^{\dagger}$ \\
        Average of two calibrated & 2.10 & 2.29 & \textbf{2.80}$^{\ast\ast}$ \\
        \method{} (2 views, calibrated) & 1.89 & \textbf{1.87}$^{\dagger}$ & \textbf{2.49}$^{\ast\ast}$ \\
        \bottomrule
    \end{tabular}
    \caption{Mean gaze-estimation error in centimeters at viewing distances of 60, 80, and 100 cm (lower is better). The lower block applies a six-point affine calibration independently for each participant and distance. Symbols report Holm-corrected paired Wilcoxon comparisons with Tobii: $^{\ast\ast\ast}$ denotes $p<0.001$, $^{\ast\ast}$ denotes $p<0.01$, and $^{\dagger}$ denotes no significant difference.}
    \label{tab:distance_evaluation}
\end{table}

\section{Conclusion and Limitations}

% We introduced \dataset{}, which captures continuous head-pose variation for fixed gaze targets, and \method{}, which uses Region Partition to fuse complementary camera evidence. Across camera count, placement, target location, and viewing distance, informative views compensate for unfavorable observations, and camera quality is more predictive than spatial spread. Region Partition directs attention toward cameras that are closer to the target and have lower single-view error, while lightweight calibration improves performance across viewing distances.

% \dataset{} uses a controlled desktop setup, fixed cameras, a limited participant group, and no glasses; \method{} also requires synchronized views. Future work will broaden data diversity and support fewer or asynchronously available cameras.

We presented \dataset{}, a multi-view dataset for screen-based gaze estimation that captures continuous head-pose variation while participants fixate on the same targets. We further proposed \method{}, a multi-view feature-fusion framework that uses region partitioning during training to encourage the integration of complementary evidence across camera views. Our evaluations across camera configurations, target locations, and viewing distances show that the main advantage of multi-view gaze estimation is its ability to compensate for unfavorable or poorly observed views using more informative viewpoints. The analysis further suggests that region partitioning promotes spatially structured attention: cameras closer to a target and with lower single-view error tend to receive greater attention during fusion. In addition, \method{} and its circle-based refinement remain effective under changes in viewing distance, especially when combined with lightweight calibration.

This work has several limitations. First, \dataset{} was collected in a controlled desktop setting with a limited number of participants, fixed camera placements, and relatively consistent illumination. Its results may therefore not fully generalize to substantially different devices, environments, demographics, or user behaviors. Second, participants removed glasses during the main data collection, so robustness to eyewear, severe occlusion, and other challenging visual conditions requires further study. Third, our method assumes synchronized multi-view inputs, which can increase hardware and deployment costs compared with single-camera solutions. Future work should investigate more diverse and longitudinal data collection, robustness to practical visual disturbances, flexible camera configurations, and efficient deployment with fewer or asynchronously available views.

\section*{Statement}

\subsection*{AI Use Statement}

In this work, we used generative AI tools to assist with the implementation of the experimental system, analysis and interpretation of experimental data, and preparation of visual materials, including tables and figures. We also used generative AI tools to assist with drafting and language editing of the manuscript. All AI-assisted outputs were reviewed, verified, and revised by the authors. The authors take full responsibility for the final content of this work, including all text, claims, analyses, and visual artifacts.

\subsection*{Ethics Statement}

This work involves the collection and planned release of synchronized facial images and gaze annotations, which may constitute sensitive biometric data. Participation was voluntary, and informed consent was obtained from all participants prior to data collection. We will release the dataset only under terms that prohibit attempts to identify participants or use the data for surveillance, discriminatory profiling, or other harmful applications. 
% We will remove direct personal identifiers and provide documentation describing the data-collection procedure, intended uses, and known limitations.
% Although the proposed method is designed for beneficial applications such as accessible and unobtrusive human--computer interaction, gaze-estimation systems can also be used to infer attention or behavior without meaningful user awareness. We therefore encourage future users to obtain appropriate consent, provide transparent disclosure, and comply with applicable privacy and data-protection regulations.

\subsection*{Reproducibility Statement}

To support reproducibility, we will release the \dataset{} dataset, source code, trained-model configurations, and scripts for preprocessing, training, and evaluation, subject to participant-consent and privacy requirements. The main paper and appendix describe the dataset protocol, subject-wise data splits, model design, and evaluation procedures. The released materials will include the configurations needed to reproduce the reported experiments and facilitate comparisons under different camera subsets, target locations, and viewing distances.

% \section*{Acknowledgments}
% Use unnumbered third level headings for the acknowledgments. All
% acknowledgments, including those to funding agencies, go at the end of the paper.

\bibliography{ref}
\bibliographystyle{iclr2027_conference}

\appendix
\section{Appendix}
\subsection{Off-Screen Target Ground Truth}
\label{apdx:off_screen_target_ground_truth}
Since both the side-view cameras and the display remain fixed throughout data collection, the mapping from screen coordinates to each side-view camera image is constant. For each side-view camera, we manually identify the four corners of the display and estimate a affine matrix between the screen coordinate system and the corresponding quadrilateral region in the camera image. The predefined 24 $\times$ 12 target grid is then projected through this affine matrix, providing an image-space ground-truth location for every on-screen fixation target.

Although the screen is visible in the side-view images, its content may reveal the currently displayed fixation target. To avoid this source of data leakage, we mask the entire screen region in the input images before they are provided to the model. This masking operation only modifies the model input and does not affect the ground-truth coordinates obtained from the geometric projection.

For off-screen targets, the physical objects remain at fixed locations throughout data collection. We manually annotate the center of each object once in each side-view camera image and use this location as its image-space ground truth. Participants are instructed to fixate on the designated object center during the corresponding trials.

\subsection{Camera Configuration Analysis}
\label{apdx:camera_configuration_analysis}

The single-camera experiments cover all eight camera positions to evaluate the baseline performance of each individual viewpoint. The two-camera experiments include all 28 pairwise combinations of the eight positions. The three-camera experiments use 18 pre-selected configurations. These configurations were selected to represent geometrically distinct layouts, including relatively compact configurations and configurations with broader spatial spread. Since the three-camera configurations do not exhaust all possible combinations, the corresponding analyses should be interpreted as exploratory.

% The eight camera positions were defined in the image plane as

% \begin{equation}
% \begin{aligned}
% &1:(0,0), \quad 2:(0,540), \quad 4:(0,1080),\\
% &5:(1920,540), \quad 6:(960,1080), \quad 7:(1920,1080),\\
% &8:(1920,0), \quad 9:(960,0).
% \end{aligned}
% \end{equation}

\subsubsection{Single-Camera Baseline Performance}

For the single-camera condition, each of the eight camera positions was trained and evaluated independently. The resulting test errors were used to rank individual camera positions. A lower rank indicates lower test error and therefore better single-camera performance.

\subsubsection{Relationship Between Single-Camera Quality and Two-Camera Performance}

To examine whether the quality of individual cameras transfers to two-camera configurations, each camera position was assigned a rank according to its single-camera test error. For a two-camera configuration consisting of cameras $i$ and $j$, we defined its mean single-camera rank as

\begin{equation}
    R_{i,j}
    =
    \frac{R_i + R_j}{2},
\end{equation}

where $R_i$ and $R_j$ are the single-camera error ranks of cameras $i$ and $j$, respectively.

We then computed the Spearman correlation between $R_{i,j}$ and the rank of the final two-camera test error across all 28 pairwise configurations. A positive correlation indicates that pairs composed of better-ranked individual cameras tend to achieve better two-camera performance.

Because the 28 two-camera configurations are constructed from only eight camera positions, they are not independent observations: each camera appears in multiple pairs, causing different configurations to share the same camera-level information. Consequently, the standard significance test for Spearman correlation, which treats pairwise observations as independent, may not provide an appropriate null distribution. We therefore use a position-label permutation test that preserves the original two-camera combinations and their test errors while breaking the association between camera position and single-camera performance.

Specifically, under the null hypothesis that single-camera performance is unrelated to the performance of two-camera configurations, we randomly reassign the eight single-camera ranks to the eight camera positions. For each permutation, we recompute the mean single-camera rank $R_{i,j}$ for all 28 camera pairs and calculate its Spearman correlation with the observed two-camera error ranking. Repeating this procedure produces a null distribution of correlation coefficients that retains the dependency structure induced by shared camera positions. The two-sided permutation $p$ value is computed as

\begin{equation}
p
=
\frac{
1 + \sum_{b=1}^{B}
\mathbb{I}
\left(
|\rho_b| \geq |\rho_{\mathrm{obs}}|
\right)
}{B+1},
\end{equation}

where $\rho_{\mathrm{obs}}$ is the observed correlation, $\rho_b$ is the correlation obtained from the $b$-th permutation, and $B$ is the number of permutations.

\subsubsection{Inter-Camera Distance in Two-Camera Configurations}

For each two-camera configuration $(i,j)$, the spatial separation between cameras was quantified by the Euclidean distance

\begin{equation}
    D_{i,j}
    =
    \sqrt{
    (x_i-x_j)^2+
    (y_i-y_j)^2
    }.
\end{equation}

We evaluated the relationship between $D_{i,j}$ and two-camera test error using both Spearman correlation and Pearson correlation. Spearman correlation tests for a monotonic relationship, whereas Pearson correlation tests for a linear relationship. Statistical significance was evaluated by randomly permuting the test errors across the 28 two-camera configurations while holding the distance values fixed.

\subsubsection{Convex-Hull Area in Three-Camera Configurations}

For each three-camera configuration, spatial coverage is quantified by the convex-hull area of the camera positions, with zero assigned to collinear layouts:

\begin{equation}
    A_{\mathrm{hull}}
    =
    \mathrm{Area}
    \left[
    \mathrm{ConvexHull}
    \left(
    \{(x_1,y_1),(x_2,y_2),(x_3,y_3)\}
    \right)
    \right].
\end{equation}

Spearman and Pearson correlations relate $A_{\mathrm{hull}}$ to test error; significance is obtained by permuting errors across the 18 configurations while keeping the convex-hull areas fixed.

\end{document}